\documentclass[twoside,11pt]{article}

\usepackage[accepted]{melba}

\usepackage{mathtools}

\usepackage{amsmath,amsfonts,bm}

\def\eqref#1{equation~\ref{#1}}

\def\1{\bm{1}}

\DeclareMathAlphabet{\mathsfit}{\encodingdefault}{\sfdefault}{m}{sl}
\SetMathAlphabet{\mathsfit}{bold}{\encodingdefault}{\sfdefault}{bx}{n}

\newcommand{\E}{\mathbb{E}}

\newcommand{\R}{\mathbb{R}}

\DeclareMathOperator*{\argmax}{arg\,max}
\DeclareMathOperator*{\argmin}{arg\,min}

\usepackage{xcolor}

 \usepackage{array}
 \usepackage{threeparttable}
 \usepackage{xcolor}
 \makeatletter
 \newcommand*{\@rowstyle}{}
\newcommand*{\rowstyle}[1]{
  \gdef\@rowstyle{#1}%
  \leavevmode\@rowstyle
  \ignorespaces
}
\newcolumntype{=}{
  >{\gdef\@rowstyle{}\ignorespaces}%
}
\newcolumntype{+}{
  >{\leavevmode\@rowstyle\ignorespaces}%
}

\usepackage{amsmath,amsfonts,amssymb}

\usepackage{lineno}   
\usepackage{amsmath}  
\usepackage{etoolbox} 

\newcommand*\linenomathpatch[1]{%
  \cspreto{#1}{\linenomath}%
  \cspreto{#1*}{\linenomath}%
  \csappto{end#1}{\endlinenomath}%
  \csappto{end#1*}{\endlinenomath}%
}

\linenomathpatch{equation}
\linenomathpatch{gather}
\linenomathpatch{multline}
\linenomathpatch{align}
\linenomathpatch{alignat}
\linenomathpatch{flalign}

\melbaheading{2022:008}{https://www.melba-journal.org/papers/2022:008.html}{2022}{1-23}{04/2022}{04/2022}{Akrami, Joshi, Aydore, Leahy} {IPMI 2021}{Aasa Feragen, Stefan Sommer, Julia Schnabel, Mads Nielsen}

\ShortHeadings{Deep Quantile Regression for Uncertainty Estimation}{Akrami et al. 2022}
\firstpageno{1}

\title{Deep Quantile Regression for Uncertainty Estimation in Unsupervised and Supervised Lesion Detection}

\author{\name Haleh Akrami \email Akrami@usc.edu \\  
	\addr Department of Biomedical Engineering, University of Southern California, Los Angeles, USA
	\AND
	\name Anand A. Joshi \email ajoshi@usc.edu \\
	\addr Department of Electrical and Computer Engineering, University of Southern California, Los Angeles, USA
	\AND
	\name Sergül Aydöre \email sergulaydore@gmail.com \\ 
	\addr Amazon Web Services, New York, USA
	\AND
	\name Richard M. Leahy \email leahy@sipi.usc.edu\\
	\addr Department of Electrical and Computer Engineering, University of Southern California, Los Angeles, USA
}

\begin{document}
%

%
%
%

%
%
%
\maketitle              

\begin{abstract}

Despite impressive state-of-the-art performance on a wide variety of machine learning tasks in multiple applications, deep learning methods can produce over-confident predictions, particularly with limited training data. Therefore, quantifying uncertainty is particularly important in critical applications such as lesion detection and clinical diagnosis, where a realistic assessment of uncertainty is essential in determining surgical margins, disease status and appropriate treatment. In this work, we propose a novel approach that uses quantile regression for quantifying aleatoric uncertainty in both supervised and unsupervised lesion detection problems. The resulting confidence intervals can be used for lesion detection and segmentation. In the unsupervised setting, we combine quantile regression with the Variational AutoEncoder (VAE). The VAE is trained on lesion-free data, so when presented with an image with a lesion, it tends to reconstruct a lesion-free version of the image. To detect the lesion, we then compare the input (lesion) and output (lesion-free) images. Here we address the problem of quantifying uncertainty in the images that are reconstructed by the VAE as the basis for principled outlier or lesion detection. The VAE models the output as a conditionally independent Gaussian characterized by its mean and variance. Unfortunately, joint optimization of both mean and variance in the VAE leads to the well-known problem of shrinkage or underestimation of variance. Here we describe an alternative Quantile-Regression VAE (QR-VAE) that avoids this variance shrinkage problem by directly estimating conditional quantiles for the input image. Using the estimated quantiles, we compute the conditional mean and variance for the input image from which we then detect outliers by thresholding at a false-discovery-rate corrected p-value.
In the supervised setting, we develop binary quantile regression (BQR) for the supervised lesion segmentation task. We show how BQR can be used to capture  uncertainty in lesion boundaries in a manner that characterizes expert disagreement.
\end{abstract}
\section{Introduction}
\label{Sec:intro}

Inference based on deep learning methods that do not take uncertainty into account can lead to over-confident predictions, particularly with limited training data \citep{reinhold2020validating}. Quantifying uncertainty is particularly important in critical applications such as clinical diagnosis, where a realistic assessment of uncertainty is important in determining disease status and appropriate treatment. For example, in the lesion detection task, knowing the uncertainty in detected boundaries may help in defining tumor margins. In the literature, predictive uncertainty is categorized into two types based on the source of uncertainty: (i) aleatoric uncertainty \citep{lakshminarayanan2016simple} that is the result of uncertainty inherent in the data, and (ii) epistemic uncertainty, which is often referred to as model uncertainty, as it is due to model limitations. Access to unlimited training data does not reduce the former uncertainty in contrast to the latter. Here we focus on aleatoric uncertainty and its estimation using quantile regression (QR) \citep{koenker1978regression}. 

A recent novel approach proposed using conditional QR to estimate aleatoric uncertainty in neural networks  \citep{romano2019conformalized,tagasovska2019single}. QR can be used to estimate the conditional median (0.5 quantiles) or other quantiles of the response variable, conditioned on the input feature variable \citep{yu2001bayesian}. QR is most commonly applied in cases where the parametric likelihood cannot be specified \citep{yu2001bayesian}, here we apply develop QR methods for Gaussian (supervised) and binary (unsupervised) applications. 

Lesion detection is an important application of deep learning in medical image processing. Here we address the important problem of learning uncertainty in order to perform statistically-informed inference for this application. Lesion detection can be applied either in a supervised framework when labels are available or with an unsupervised framework using a generative model such as the VAE. We describe how aleatoric uncertainty can be quantified in both of these settings using quantile regression to define confidence intervals, which are then used to identify lesions. In both supervised and unsupervised frameworks, we apply quantile regression by changing the loss function of the network. For quantile regression in the unsupervised setting, we use the formulation developed by \cite{he1997quantile}.  Our goal is to learn the characteristics of the input distribution to separate inliers from outliers. The quantiles help us to define  confidence intervals from which we can identify outliers. For the supervised setting we use binary quantile regression \citep{koenker2001quantile} in order to capture uncertainty in binary classification problems.  In both scenarios, our goal is to estimate aleatoric uncertainty in segmentation and calculate a confidence interval associated with each segmentation.

\textbf{Unsupervised Lesion Detection:} Generative models, including  autoencoders, can be used for unsupervised lesion detection. Once the distribution of anomaly-free samples is learned during the training, during inference we can compute the reconstruction error between a given image and its reconstruction to identify abnormalities \citep{aggarwal2015outlier,akrami2021quantile,akrami2020brain,akrami2019robust}. Decisions on the presence of outliers are often performed based on empirically chosen thresholds. Here we use quantile regression to define a principled-approach to thresholding.

Variational autoencoder (VAE) \citep{kingma2013auto} and its variants can approximate the underlying distribution of high-dimensional data. VAEs are trained using the variational lower bound of the marginal likelihood of data as the objective function. They can then be used to generate samples from the data distribution, where probabilities at the output are modeled as parametric distributions such as Gaussian or Bernoulli that are conditionally independent across output dimensions \citep{kingma2013auto}. By using VAE,  An and Cho \citeyearpar{an2015variational} proposed to use reconstruction probability rather than the reconstruction error to detect outliers. This allows a more principled approach to anomaly detection since inference is based on quantitative statistical measures and can include corrections for multiple comparisons.   

For determining the reconstruction probability, we need to predict both conditional mean and variance using VAEs for each of the output dimensions. The estimated variance represents an aleatoric uncertainty associated with the  conditional estimates given the data \citep{reinhold2020validating}. Estimating variance is more challenging than estimating the mean in generative networks due to the unbounded likelihood \citep{skafte2019reliable}. In the case of VAEs, if the conditional mean network prediction is nearly perfect (zero reconstruction error), then maximizing the log-likelihood pushes the estimated variance towards zero in order to maximize the likelihood. This also makes VAEs susceptible to overfitting the training data giving a near-perfect reconstruction on the training data and very small uncertainty. This near-zero variance does not reflect the true performance of the VAE on the test data. For this reason, near-zero variance estimates, with the log-likelihood approaching an infinite supremum, do not lead to a good generative model. It has been shown that there is a strong link between this likelihood blow-up and the mode-collapse phenomenon \citep{mattei2018leveraging,reinhold2020validating}. In fact, in this case, the VAE behaves much like a deterministic autoencoder \citep{blaauw2016modeling}. 

While the classical formulation of VAEs allows both mean and variance estimates \citep{kingma2013auto}, because of the variance shrinkage problem, most if not all implementations of VAE, including the standard implementations in PyTorch and Tensorflow libraries, estimate only the mean with a fixed value of variance \citep{skafte2019reliable}. Here we describe an approach that overcomes the variance shrinkage problem in VAEs using quantile regression (QR) in place of variance estimation. We then demonstrate the application of this new QR-VAE by computing reconstruction probabilities for a brain lesion detection task.

\textbf{Supervised Lesion Detection:} Labelled training data are preferable, if available, as they lead to better performance compared to unsupervised models \citep{you2019unsupervised}. One approach to estimating  uncertainty is to use the soft-max probability of the cross-entropy loss  \citep{devries2018leveraging}. Softmax probabilities are known to be poorly calibrated, and imperceptible perturbations to the input image can change the deep network’s softmax output significantly \citep{guo2017calibration}. Softmax confidence also conflates two different sources of uncertainty (aleatoric and epistemic). Bayesian neural networks \citep{neal2012bayesian} can be used for estimating aleatoric uncertainty by measuring conditional entropy; however, these models are not able to capture multimodal uncertainty profiles \citep{tagasovska2019single}. An alternative method to capture aleatoric uncertainty is using quantile regression \citep{tagasovska2018single}. Here we used binary quantile regression 
\citep{kordas2006smoothed,manski1985semiparametric} to capture the quantiles of labels which can be used to define multiple nested segmentation masks with increasing uncertainty. Our goal is to capture the source of uncertainty within the data distribution where there is more than one plausible answer to the segmentation problem due to disagreement between the specialists who labeled the data. Binary quantile regression is also robust to label noise \citep{oh2016bayesian}. Finally, Kordas et al. \citeyearpar{kordas2006smoothed} showed that binary quantile regression can be useful for unbalanced data and leads to a more comprehensive view on how the predictor variables influence the response.

\textbf{Related Work:}
A few recent papers have targeted the variance shrinkage problem. Among these, \cite{detlefsen2019reliable} describe reliable estimation of the variance using Comb-VAE, a locally aware mini-batching framework that includes a scheme for unbiased weight updates for the variance network. In an alternative approach Stirn and Knowles 2020, \citep{stirn2020variational}
suggest treating variance variationally, assuming a Student’s $t$-likelihood for the posterior to prevent optimization instabilities and a Gamma prior for the precision parameter of this distribution. 
The resulting
Kullback–Leibler (KL) divergence induces gradients that prevent the
variance from approaching zero \citep{stirn2020variational}. 

In the supervised framework, several papers estimate uncertainty for segmentation; however, only a few separately consider aleatoric uncertainty and focus on multi-rated labels \citep{czolbe2021segmentation,hu2019supervised,islam2021spatially,kohl2018probabilistic}. Czolbe et al. \citeyearpar{czolbe2021segmentation} compared these methods to investigate whether they are helpful in an assessment of segmentation quality and active learning. Recently, Monteiro et al. \citeyearpar{monteiro2020stochastic} used a stochastic segmentation network for modeling spatially correlated uncertainty in
image segmentation. They
applied a multivariate Normal distribution over the softmax
logits and used low-rank approximation to estimate the full covariance matrix across all
the pixels in the image \citep{monteiro2020stochastic}.

\textbf{Our Contribution:} In the unsupervised setting, to obtain a probabilistic threshold and address the variance shrinkage problem, we suggest an alternative and attractively simple solution. Assuming the output of the VAE has a Gaussian distribution, we quantify uncertainty in VAE estimates using conditional quantile regression (QR-VAE). The aim of conditional quantile regression \citep{koenker1978regression} is to estimate a quantile of interest. Here we use these quantiles to compute variance, thus sidestepping the shrinkage problem. It has been shown that quantile regression is able to capture aleatoric uncertainty  \citep{tagasovska2019single}. We demonstrate the effectiveness of our method quantitatively and qualitatively on simulated and brain MRI datasets. Our approach is computationally efficient and does not add any complications to training or sampling procedures. In contrast to the VAE loss function, the QR-VAE loss function does not have an interaction term between quantiles, and therefore, shrinkage does not happen. Since quantile regression does not satisfy finite-sample coverage guarantees, we applied conformalized quantile regression \citep{romano2019conformalized}, which combines conformal prediction with classical quantile regression, to have the theoretical guarantee of valid
coverage.

We also use binary quantile regression in a supervised framework in order to capture the uncertainty of lesion annotations. We demonstrate estimation of multiple quantiles in imaging data in which each lesion is delineated by four human observers and compare to human-rater ground truth and a binary cross-entropy formulation.    

A preliminary version of these results was presented in \cite{akrami2021quantile}. The novel extensions presented in the current work include: (1) application of conformalized quantile regression to unsupervised learning (section \ref{subsec:qr-vae}); (2) extension of unsupervised approach to the supervised approach using binary quantile regression (section \ref{subsec:bqr-U-Net}), (3) application of binary quantile regression for lesion detection and uncertainty estimation, (section \ref{subsec:sup-lesion-detection}), and (4) additional results and validation that extend those in the earlier paper.  We provide a public version of our code at \url{https://github.com/ajoshiusc/QRSegment} and \url{https://github.com/ajoshiusc/QRVAE}. 

\section{Background}

\subsection{Variance Shrinkage Problem in Variational Autoencoders}
Let $x_{i} \in \mathbb{R}^D$ be an observed sample of random variable ${X}$ where $i \in \{1, \cdots, N \}$, $D$ is the number of features and $N$ is the number of samples; and let ${z_{i}}$ be an observed sample for latent variable ${Z}$. Given a sample $x_{i}$ representing the input data, the VAE is a probabilistic graphical model that estimates the posterior distribution $p_{\theta}({Z}|{X})$ as well as the model evidence $p_{\theta}({X})$, where $\theta$ are the generative model parameters \citep{kingma2013auto}. The VAE approximates the posterior distribution of ${Z}$ given ${X}$ by a tractable parametric distribution and minimizes the ELBO (evidence lower bound) loss \citep{an2015variational}. It consists of  the encoder network that computes $q_{\phi}(Z|X)$,
and the decoder network that computes $p_{\theta}(X|Z)$ \citep{wingate2013automated}, where ${\phi}$ and ${\theta}$ are model parameters. Since we use the neural network for learning the distributions $q_{\phi}(Z|X)$ and $p_{\theta}(X|Z)$, the parameters $\theta$ and $\phi$ are modeled by the weights of the encoder and decoder networks and will be learned from the data during training. The marginal likelihood of an individual data point can be rewritten as follows:

\begin{align}
\begin{split}
\log p_{\theta}(x_{i}) = D_{KL}(q_{\phi}(Z|x_{i}),p_{\theta}(Z|x_{i}))
  + L(\theta,\phi;x_{i}),
\end{split}
\end{align}
where
\begin{align}
\begin{split}
    L(\theta,\phi;x_{i})= \E_{q_{\phi}(Z|x_{i})}[\log(p_{\theta}(x_{i}|Z))] - D_{KL}(q_{\phi}(Z|x_{i})||p_{\theta}(Z)).
\end{split}
\label{eqn:overall_loss}
\end{align}
The first term (log-likelihood) in equation \ref{eqn:overall_loss} can be interpreted as the  \emph{reconstruction loss} and the
second term (KL divergence) as the \emph{regularizer}. The total loss over all samples can be written as:
\begin{align}
L({\theta},{\phi},X) &= L_{REC}+L_{KL}
\end{align}
where $L_{REC} \coloneqq \mathbb{E}_{q_{\phi}({Z|X})}[\log(p_{\theta}({X}|{Z}))]$
and $ L_{KL} \coloneqq  D_{KL}(q_{\phi}({Z|X})||p_{\theta}({Z}))$.

Assuming the posterior distribution is Gaussian and using a 1-sample approximation \citep{skafte2019reliable}, the likelihood term  simplifies to:
\begin{align}
L_{REC}=\sum_i\frac{-1}{2}\log(\sigma_{\theta}^{2}(z_{i}))-\frac{(x_{i}-\mu_{\theta}(z_{i}))^{2}}{2\sigma_{\theta}^{2}(z_{i})}
\end{align}
where $ Z \sim p(Z) =  {N}(0,\,I)$ ($I$ is identity matrix), $X|Z \sim p_{\theta}(X|Z)=  {N}(X|\mu_{\theta}(Z),\,\sigma_{\theta}(Z))$, and
$Z|X \sim q_{\phi}(Z|X)=  {N}(Z|\mu_{\phi}(X),\,\sigma_{\phi}(X))$.
$\mu_{\theta}(Z), \sigma_{\theta}(Z)$ are posterior mean and variance; $\mu_{\phi}(X)$, and $\sigma_{\phi}(X)$ are encoder mean and variance. 
Optimizing VAEs over mean and variance with a Gaussian posterior is challenging \citep{skafte2019reliable}. If the model has sufficient capacity that there exists $(\phi,\theta)$ for which $\mu_{\theta}(z)$ provides a sufficiently good reconstruction, then the second term pushes the variance to zero before the term
$\frac{-1}{2}\log(\sigma_{\theta}^{2}(x_{i})))$ catches up \citep{blaauw2016modeling,skafte2019reliable}. 

One practical example of this behavior is in speech processing applications \citep{blaauw2016modeling}. The input is a spectral envelope which is a relatively smooth 1D curve.  Representing this as a 2D image produces highly structured and simple training images. As a result, the model  quickly learns how to accurately reconstruct the input. Consequently, reconstruction errors are small and the estimated variance becomes vanishingly small. 
Another example is 2D reconstruction of MRI images where the images from neighbouring 2D slices are highly correlated leading again to variance shrinkage \citep{volokitin2020modelling}. 
To overcome this problem, variance estimation networks can be avoided using a Bernoulli distribution or  by simply setting variance to a constant value \citep{skafte2019reliable}. 

\subsection{Conditional Quantile Regression }
In contrast to classical parameter estimation where the goal is to estimate the conditional mean of the response variable given the feature variable, the goal of quantile regression is to estimate conditional quantiles based on the data  \citep{yu2001bayesian}. 
The most common application of quantile regression models is in cases in which a parametric likelihood cannot be specified \citep{yu2001bayesian}. Another motivation for quantile regression is that quantiles are robust to outliers \citep{john2015robustness}.

Quantile regression can be used to estimate the conditional median (0.5 quantile) or other quantiles of the response variable conditioned on the input data. The $\alpha$-th conditional quantile function is defined as $q_{\alpha}(x) \coloneqq \inf\{y\in\R:F(y|X=x)\geq \alpha\}$ where $F=P(Y \leq y)$ is a strictly monotonic cumulative distribution function. Similar to classical regression analysis which estimates the conditional mean, the $\alpha$-th quantile regression $(0<\alpha<1)$ seeks a solution to the following minimization problem for input $x$ and output $y$ \citep{koenker1978regression,yu2001bayesian}:
\begin{align}
\argmin \limits_{\theta}\sum_{i}\rho_{\alpha}(y_{i}-f_{\theta}(x_{i}))
\label{q_loss}
\end{align}
where $x_i$ are the inputs, $y_i$ are the responses,  $\rho_{\alpha}$ is the \emph{check function} or \emph{pinball loss} \citep{koenker1978regression} and $f$ is the model paramaterized by $\theta$. The goal is to estimate the parameter $\theta$ of the model $f$. The  \emph{pinball loss} is defined as:

\begin{align}
\rho_{\alpha}(y,f_{\theta}(x_{i})):=\begin{cases}
            \alpha(y-f_{\theta}(x_{i}))   &\text{if $(y-f_{\theta}(x_{i})) > 0$ }  \\
            (1-\alpha)(y-f_{\theta}(x_{i}))   &\text{otherwise.}
        \end{cases}
\end{align}
Due to its simplicity and generality, quantile regression is widely applicable in
classical regression and machine learning to obtain a conditional
prediction interval \citep{rodrigues2020beyond}.
It can be shown that minimization of the loss function in equation \ref{q_loss}  is  equivalent to maximization of the likelihood function formed by combining independently distributed asymmetric Laplace densities \citep{yu2001bayesian}:
\begin{align*}
\argmax \limits_{\theta} L(\theta)=\frac{\alpha(1-\alpha)}{\sigma} \exp\left\{\frac{-\sum_i \rho_{\alpha}(y_{i}-f_{\theta}(x_{i}))}{\sigma}\right\}
\end{align*}
where $\sigma$ is the scale parameter. 
 Individual quantiles can be shown to be maximum likelihood estimates of Laplacian density. In this paper we estimate two quantiles jointly and therefore our loss function can be seen as a summation of two Laplacian likelihoods.

\section{Deep Uncertainty Estimation with Quantile Regression}
\subsection{Quantile Regression Varational Autoencoder (QR-VAE)}
\label{subsec:qr-vae}
Instead of estimating the conditional mean and conditional variance directly at each pixel (or feature), the outputs  of our QR-VAE are multiple quantiles of the output distributions at each pixel. This is achieved by replacing the Gaussian likelihood term in the VAE loss function with the quantile loss (check or pinball loss). For the QR-VAE, if we assume a Gaussian output, then only two quantiles are needed to fully characterize the Gaussian distribution. Specifically, we estimate the median and $0.15$-th quantile, which corresponds to approximately one standard deviation (more precisely 1.036 std dev.) from the mean under the Gaussian model. Our QR-VAE ouputs, $Q_{L}$ (low quantile) and $Q_{H}$ (high quantile), are then used to calculate the mean and the variance. To find these conditional quantiles, fitting is achieved by
minimization of the pinball loss for each quantile. The resulting reconstruction loss for the proposed model can be calculated as:
\begin{align*}
L_{REC}=\sum_{i}\rho_{L}(x_{i}-f_{\theta_L}(x_{i}))+\sum\rho_{H}(x_{i}-f_{\theta_H}(x_{i}))
\end{align*}
where $\theta_L$ and $\theta_H$ are the parameters of the models corresponding to the quantiles $Q_L$ and $Q_H$, respectively.
The minimization of this loss results in the desired quantile estimates for each output pixel.

We reduce the chance of quantile crossing (consistency in the quantiles defined by: $Q_{\tau_1} \subset Q_{\tau_2}$ when $\tau_1 > \tau_2$) by limiting the flexibility  of independent quantile regression. This is done by simultaneous estimation of both quantiles with one neural network, rather than training separate networks for each quantile \citep{rodrigues2020beyond}. Note that the estimated quantiles share network parameters except for the last layer.

While quantile regression guarantees coverage of data (based on the quantiles chosen) in the training set, performance on a held-out validation data is not guaranteed. 
In order to have a coverage guarantee for finite samples on unseen data, we deployed conformalized quantile regression using a calibration set as explained in \cite{romano2019conformalized}. 
Conformal predictions provide a non-asymptotic, distribution-free coverage guarantee \citep{shafer2008tutorial}. The main idea of conformal prediction is to fit a model on the training data, then use the residuals on held-out calibration data to quantify the uncertainty in future predictions. This offers finite sample distribution-free performance guarantees.
The conformalized quantile regression approach combines conformal prediction with quantile regression \citep{sesia2020comparison}. For this, we use the approach presented in \cite{romano2019conformalized} that combines conformal prediction with quantile regression. This approach inherits both the finite sample, distribution-free
validity of conformal prediction and the statistical efficiency of quantile regression.

\subsection{Binary Quantile Regression U-Net (BQR U-Net)}
\label{subsec:bqr-U-Net}
For calculating quantiles of a binary response, consider the following model:
\begin{align*}
Y^*=h(x)+\epsilon,\ \ \ Y=I\{Y^* \geq 0\}
\end{align*}
where $Y^*$ is a hidden variable, $h(x)$ is the true model, and $\epsilon$ is the noise. No distribution $\epsilon$ is assumed for the model \citep{kordas2006smoothed}.
Since the indicator function is monotone increasing, and since quantiles are invariant under monotone transform,  we have:
 \begin{align*}
Q_\tau(Y|X)=I(Q_\tau(Y^*|X)).
\end{align*} 
$Q_{\tau}(Y|X)$ is the $\tau th$ conditional quantile of $Y$ given $X$. By modeling $Q_\tau(Y^*|X) = f_\tau(X,\beta)$ with $\beta$ as the parameter, the $\beta$ parameter can be estimated by:

\begin{align*}
\argmin \limits_{\beta}\sum_{i}\rho_{\tau}(y_{i}-I(f_\tau(x_{i},\beta)\geq 0))
\end{align*}

\noindent
which can be shown to be
equivalent to a maximization problem \citep{kordas2006smoothed}:
\begin{align*}
\argmax \limits_{\beta}\sum_{i}[y_{i}-(1-\tau)]I(f_\tau(x_{i},\beta)\geq 0)
\end{align*}
However, the function is not differentiable, because of the
use of the indicator function. To apply gradient based
optimization methods for training the neural network, we use the smoothed
approximation \citep{kordas2006smoothed}:
\begin{align}
\argmax \limits_{\beta}\sum_{i}[y_{i}-(1-\tau)]K(f_\tau(x_{i},\beta))
\end{align}
where $K(t)$ is smoothed version of the
indicator function, with the following properties:
\begin{align*}
K(t)\geq 0,\, \forall t \in R, \lim_{t \to +\infty} K(t)=1, \lim_{t \to -\infty}K(t)=0.
\end{align*} Specifically, to train the neural networks, we choose $K(t)= \frac{1}{1+e^{-t}}$, the sigmoid function, which has the desired properties.

In this paper we us the BQR loss to solve the lesion detection and segmentation task. We use a U-Net architecture with multiple heads (output branches), where each head estimates a specific quantile for the labels at the pixel level. We observed that joint estimation of multiple quantiles is computationally faster than solving for each separately, and also avoids the quantile crossing problem. 

A standard U-Net would use a cross-entropy loss for this segmentation task. Here we replace this with the BQR loss. To estimate the $n$-th quantile, the BQR loss is given by:
\begin{align}
\label{eq:bqr_loss}
\text{Loss} = \sum_n\sum_{i}[y_{i}-(1-\tau_n)]K(f_{\tau_{n}}(x_{i},\beta_n))
\end{align}
where each $f$ correspond to a head of U-Net, $\tau_1 ... \tau_n$ shows different quantiles and $\beta_1 ... \beta_n$ are estimated parameter for each quantile respectively. A single network is used to estimate all quantiles. We chose to train a single network with output branches for each quantile since there is a common network across quantiles except for the last layer. This makes training of the quantiles consistent avoiding crossing of the estimated quantiles. We choose $K(t)$ to be the sigmoid function.
\section{Experiments and Results}
We evaluate our proposed approaches for supervised and unsupervised deep quantile regression on (i) A simulated dataset for density estimation, (ii) Unsupervised lesion detection in a brain imaging dataset, and (iii) Supervised lesion detection in a lung cancer dataset.  For the simulated data, we compare our results qualitatively and quantitatively, using KL divergence between the learned distribution and the original distribution, with Comb-VAE \citep{skafte2019reliable} and VAE as baselines. For unsupervised lesion detection, we compare our lesion detection results with the VAE, which estimates both mean and variance. The area under the receiver operating characteristic curve (AUC) and dice coefficients are used as performance metrics. We also performed the unsupervised lesion detection task nonparametrically, estimating upper and lower quantiles of the images and then assigning voxel lesion labels if their intensities are outside those quantiles. Using the BQR U-Net, we estimated the thresholded probability of the labels for a dataset with multiple (4) annotators per image. We compared the dice coefficient of these thresholded areas obtained using the BQR U-Net with their corresponding counterparts calculated both using the softmax probability of a deterministic U-Net and the ground truth (as determined by the four human raters).

\subsection{Simulations for VAE}
Following \cite{skafte2019reliable}, we first evaluate variance estimation using VAE, Comb-VAE, and QR-VAE on a simulated dataset. The two moon dataset inspires the data generation process for this simulation\footnote{\url{https://scikit-learn.org/stable/modules/generated/sklearn.datasets.make_moons}}. First, we generate 500 points in $\mathbb{R}^{2}$ in a two-moon manner to generate a known two-dimensional latent space. These are then mapped to four dimensions  ($v_{1}$, $v_{2}$, $v_{3}$, $v_{4}$) using the following equations:
\begin{align*}
v_{1}(z_{1},z_{2})=z_{1}-z_{2}+ \epsilon \sqrt {0.03+0.05(3+z_{1})}\\
v_{2}(z_{1},z_{2})=z_{1}^{2}-\frac{1}{2}z_{2}+ \epsilon \sqrt {0.03+0.03||{z_{1}}||_{2}}\\
v_{3}(z_{1},z_{2})=z_{1}z_{2}-z_{1}+ \epsilon \sqrt {0.03+0.05||{z_{1}}||_{2}}\\
v_{4}(z_{1},z_{2})=z_{1}+z_{2}+ \epsilon \sqrt {0.03+\frac{0.03}{0.02+||{z_{1}}||_{2}}}
\end{align*}
where $\epsilon$ is sampled from a normal distribution. For more details about the simulation, please refer to \cite{skafte2019reliable}\footnote{\url{https://github.com/SkafteNicki/john/blob/master/toy_vae.py}}. After training the models, we first sample from $z$ using a Gaussian prior, and input that sample to the decoder to generate parameters of the posterior $p_\theta(x|z)$, and then sample again from the posteriors using the estimated means and variances from the decoder. The distribution of these generated samples represents the learned distribution in the generative model. 

\begin{figure}
\centering
  \includegraphics[width=0.8\textwidth]{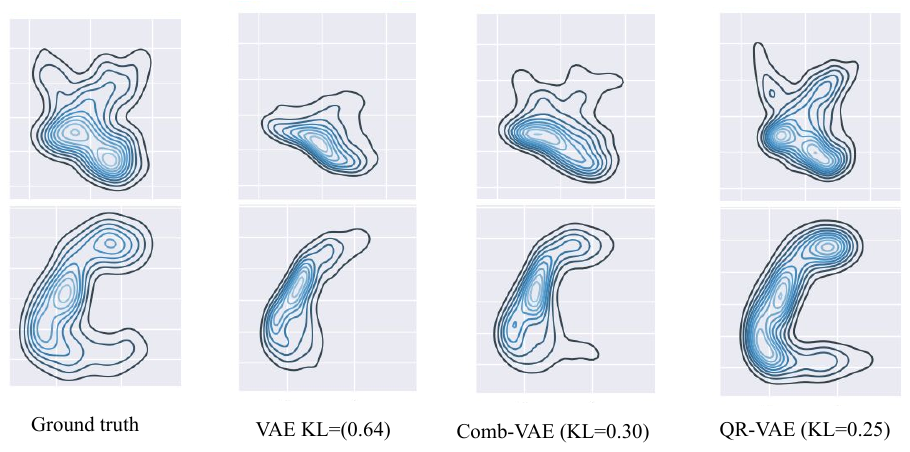}
  \caption{Pairwise joint distribution of the ground truth and generated distributions.
Top: $v_{1}$ vs. $v_{2}$ dimensions. Bottom: $v_{2}$ vs $v_{3}$ dimensions. From left to right: original distribution and distributions computed using VAE, Comb-VAE and QR-VAE, respectively. We also list the KL divergence between the learned distribution and the original distribution in each case.}
  \label{fig:prob_dist}

\end{figure}

In Figure \ref{fig:prob_dist}, we plot the pairwise joint distribution for the input data as well as the generated samples using various models. We used Gaussian kernel density estimation  to model the distributions from 1000 samples in each case. We observe that the standard VAE underestimates the variance resulting in insufficient learning of the data distribution. The samples from our QR-VAE model capture a data distribution more similar to the ground truth than either standard VAE or Comb-VAE. Our model also outperforms VAE and Comb-VAE in terms of KL divergence between input samples and generated samples as can be seen in Figure 1. The KL-divergence is calculated using universal-divergence, which estimates the KL divergence based on k-nearest-neighbor (k-NN) distance \citep{wang2009divergence}\footnote{\url{https://pypi.org/project/universal-divergence}}. 

\subsection{Unsupervised Lesion Detection}
\subsubsection{Network Architecture}
Next, we investigate utility of the proposed QR-VAE for the medical imaging application of detecting brain lesions. 
Multiple automatic lesion detection approaches have been developed to assist clinicians in identifying and delineating lesions caused by congenital malformations, tumors, stroke or brain injury. The VAE is a popular framework among the class of unsupervised methods \citep{chen2018unsupervised,baur2018deep,pawlowski2018unsupervised}. After training a VAE on a lesion free dataset, presentation of a lesioned brain to the VAE will typically result in reconstruction of a lesion-free equivalent. The error between input and output images can therefore be used to detect and localize lesions. However, selecting an appropriate threshold that differentiates lesion from noise is a difficult task. Furthermore, using a single global threshold across the entire image will inevitably lead to a poor trade-off between true and false positive rates. 
Using the QR-VAE,  we can compute the conditional mean and variance of each output pixel. This allows a more reliable and statistically principled approach for detecting anomalies by thresholding based on computed $p$-values. Further, this approach also allows us to correct for multiple comparisons. 

The network architectures of the VAE and QR-VAE are chosen based on previously established architectures \citep{larsen2015autoencoding}. Both the VAE and QR-VAE consist of three consecutive blocks of convolutional layer, a batch normalization
layer, a rectified linear unit (ReLU) activation
function and a fully-connected layer in the bottleneck
for the encoder. The decoder includes  three consecutive blocks
of deconvolutional layers, a batch normalization layer
and ReLU followed by the output layer that has 2 separate deconvolution
layers (for each output) with sigmoid activations. For the VAE, the outputs represent mean and variance while for QR-VAE the outputs represent two quantiles from which the conditional mean and variance are computed at each voxel.
The size of the input layer is $3 \times 64 \times 64$ where the first dimension represents three different MRI contrasts:  T1-weighted, T2-weighted, and FLAIR for each image. 

\subsubsection{Training, Validation, and Testing Data}
For training we use 20 central axial slices of brain MRI datasets 
from a combination of 119 subjects from the Maryland MagNeTS study 
\citep{gullapalli2011investigation} of neurotrauma and 112 subjects from the TrackTBI-Pilot 
\citep{yue2013transforming} dataset, both available for download from  
\url{https://fitbir.nih.gov}.
We use 2D slices rather than 3D images to make sure we had a large enough dataset for training the VAE. 
These datasets contain T1, T2, and FLAIR images for each
subject, and have sparse lesions. We have found that in practice both VAEs have some robustness to lesions in these training data so that they are sufficient for the network to learn to reconstruct lesion-free images as required for our anomaly detection task. The three imaging modalities  (T1, T2, FLAIR) were rigidly co-registered within subject and to the MNI brain atlas reference and re-sampled to 1mm isotropic resolution. Skull and other non-brain tissue were removed using BrainSuite (\url{https://brainsuite.org}). Subsequently, we reshaped each sample into $64 \times 64$ dimensional images and performed histogram equalization to a lesion free reference.  We separated 40 subjects as the validation/calibration set.

We evaluated the performance of our model on a test set consisting of 20 central axial slices of 28 subjects from the ISLES (The Ischemic Stroke Lesion Segmentation) database \citep{maier2017isles} for which ground truth, in the form of manually-segmented lesions, is also provided. We performed similar pre-processing as for the training set.
\begin{figure}
\centering
  \includegraphics[width=0.65\textwidth]{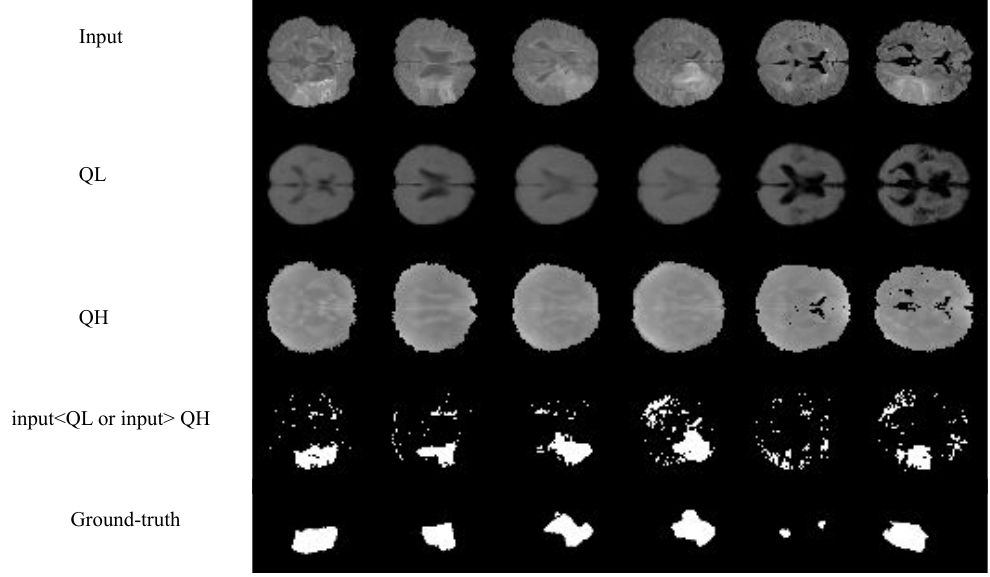}
  \caption{Model-free lesion detection for ISLES dataset using $Q_L=Q_{0.025}$ and $Q_H=Q_{0.975}$. Pixels outside the [$Q_L$, $Q_H$] interval are marked outliers. Estimated quantiles are the outputs of QR-VAE.}
  \label{fig:isles_results2}
\end{figure}
\begin{figure}
\centering
  \includegraphics[width=1\textwidth]{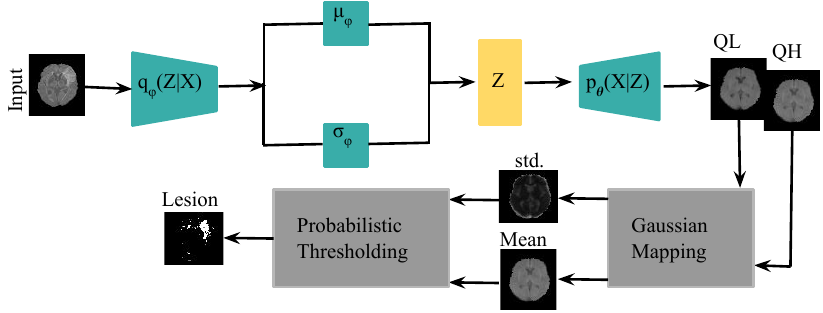}
  \caption{Estimating two quantiles in the ISLES dataset using QR-VAE. Using the Gaussian assumption for the posterior, there is 1-1 mapping from these quantiles to mean and standard deviation.}
  \label{fig:anoma_det2}
\end{figure}

\begin{figure}
\begin{center}
\includegraphics[width=0.8\textwidth]{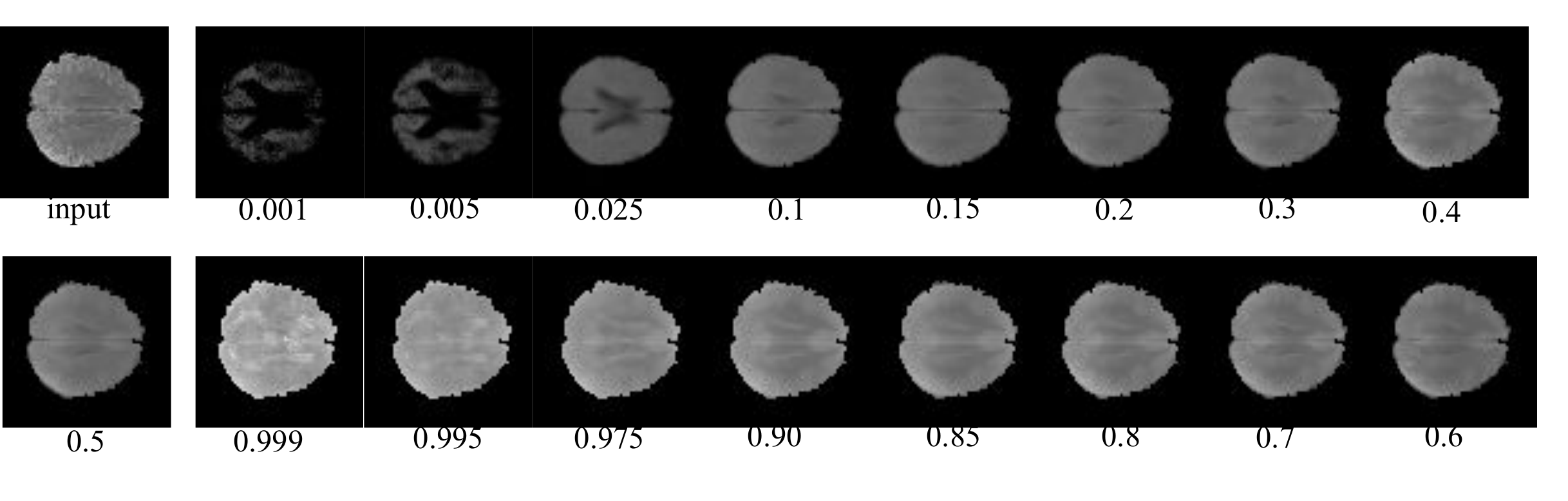}
\end{center}

\caption{Pixel-wise quantile image thresholds for a single test image as
a function of quantile computed using the QR-VAE.}
\label{fig:multi_QR}
\end{figure}

\begin{figure}
\begin{center}
\includegraphics[width=0.8\textwidth]{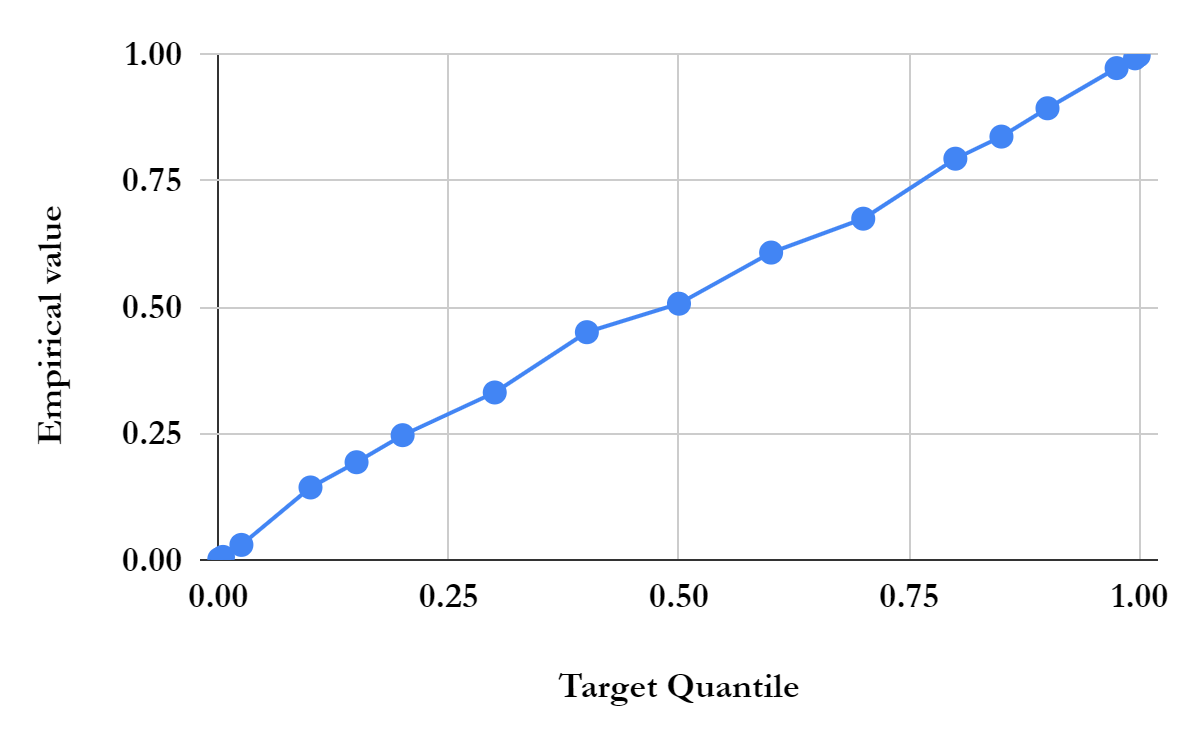}
\end{center}

\caption{Vertical axis indicates the fraction of pixels in the entire
testing set whose intensity is below the corresponding quantile for that
pixel as computed using the QR-VAE.  Note that as aggregated over the
entire test set, the computed pixel-wise quantiles closely match the
true distribution assuming anomaly-free data (in practice the fraction
of anomalous pixels is a very small fraction of the total, so the
presence of lesions in the data should not substantially affect this
plot).}
\label{fig:multi_QR2}
\end{figure}

\subsubsection{Model-free Anomaly Detection}
For simplicity, we first performed the lesion detection task using the QR-VAE without the Gaussian assumption as shown in Figure \ref{fig:isles_results2}. We trained the QR-VAE to estimate the $Q_{0.025}$ and $Q_{0.975}$ quantiles. We then used these quantiles directly to threshold the input images for anomalies. This leads to a nominal 5\% (per pixel) false positive rate. This method is simple
and avoids the need for validation data to determine an appropriate threshold. However,
without access to $p$-values we are unable to determine a threshold that can be used to correct
for multiple comparisons by controlling the false-discovery or family-wise error rate. A model needs to be assumed to obtain $p$-value as we do in  \ref{subsubsec-gaussian-model-anomaly}. The Dice coefficient for this model was 0.37 and 0.32 with and without conformalization respectively (see Table \ref{tab:1}). 
For validating the accuracy of the computed quantiles we calculated the the percentage of pixels that lie below the estimated quantiles. Even in the extreme quantiles the percentage of pixels with lower intensity than the threshold predicted by each quantile was very close to each of the estimated quantile values (Figures  \ref{fig:multi_QR} and \ref{fig:multi_QR2}).

\subsubsection{Gaussian model for anomaly detection} 
\label{subsubsec-gaussian-model-anomaly}
In a second experiment, we trained a VAE with a Gaussian posterior and the QR-VAE as illustrated in Figure \ref{fig:anoma_det2}, in both cases estimating conditional mean and variance. Specifically, we estimated the $Q_{0.15}$ and $Q_{0.5}$ quantiles for the QR-VAE and used these values to compute pixel-wise mean and variance assuming a Gaussian model. By comparing the pixel intensity to the Gaussian model values, we can compute $p$-values for each pixel.
In order to identify or segment lesions, we threshold the pixel-wise $p-$values. Naively applying a threshold separately at each pixel will result in a large number of false positives because of the multiple comparisons problem \citep{shaffer1995multiple}. For example, if all pixels are independent, and follow the null distribution, then thresholding at an $\alpha =0.05$ significance level value would lead to 5\% of all pixels being identified as lesion, even though none were present. While in practice this number is much lower because of spatial correlation in the image, it is still important to account for multiple comparisons. 

\begin{figure}
\centering
\includegraphics[width=0.64\textwidth]{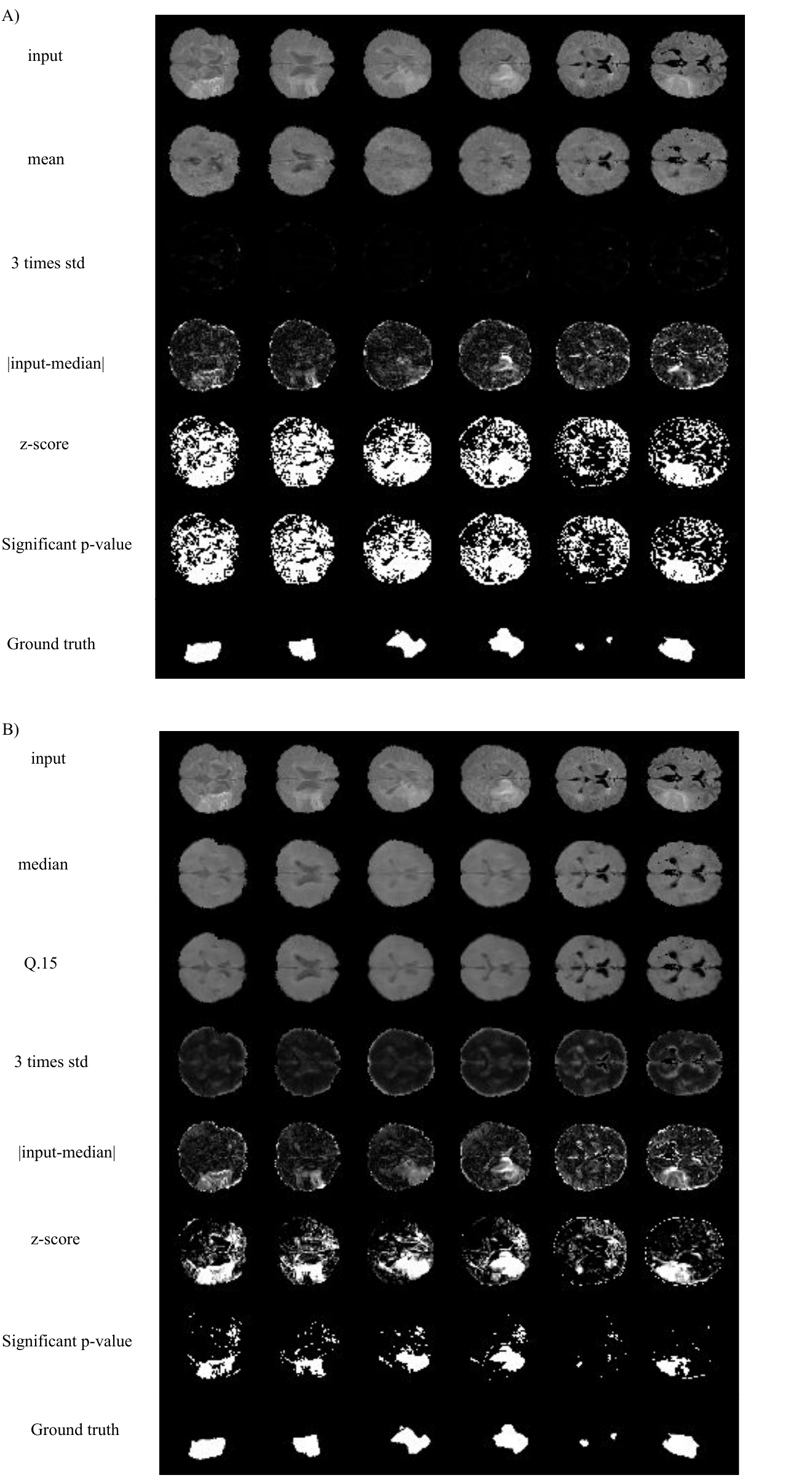}
\caption{Lesion detection for the ISLES dataset. A) VAE with mean and variance estimation  B) QR-VAE.
First, we normalize the error value using the pixel-wise model's estimates of mean and variance. The resulting z-score is then converted to an FDR-corrected $p$-value and the images are thresholded at a significance level of 0.05. The bottom rows represent ground truth based on expert manual segmentation of lesions.}
\label{fig:isles_results1}
\end{figure}

The best known such adjustment is the Bonferroni correction \citep{bland1995multiple}. In medical imaging applications, this correction tends to be too conservative since
pixels are correlated. Other methods for multiple comparison correction are designed to control the Family-Wise Error Rate (FWER, probability
of making one or more false discoveries) \citep{tukey1953problem} or the False Discovery Rate (FDR) \citep{benjamini1995controlling}. The FDR is the expected ratio of the number of false positives to the total number of positives (rejections of the null). In other words, in an FDR-corrected thresholding at an $\alpha =0.05$ significance level, we would expect 5\% of the detected lesion pixels to be false positives. Here we use the 
Benjamini-Hochberg procedure \citep{benjamini1995controlling} with $\alpha =0.05$. 
As shown in Figure \ref{fig:isles_results1}, the VAE underestimates the variance, so that most of the brain shows significant $p$-values, even with FDR correction. On the other hand, the QR-VAE's thresholded results detect anomalies that reasonably match the ground truth. 
To produce a quantitative measure of performance, we also computed the area under the ROC curve (AUC) for VAE and QR-VAE. To do this we first computed z-score images by subtracting the mean and normalizing by standard deviation. We then applied a median filtering with a $7\times 7$ window. By varying the threshold on the resulting images and comparing it to ground truth, we obtained AUC values of 0.52 for the VAE and 0.92 for the QR-VAE. We also obtained Dice coefficient values of 0.006 for the VAE and 0.37 for the QR-VAE. All quantitative measurements were computed on a voxel-wise basis. Also note that with the conformalized formulation, the Dice coefficient further increased to 0.41 (Table \ref{tab:1}). Results using the conformalized formulation improved performance of the QR-models by calibrating the quantiles, resulting in an increase in the Dice coefficients in both Gaussian and model-free cases.

\begin{table}

\caption{Compariaon of the performance of unsupervised lesion detection for VAE and QR-VAE, with and without conformalization. QR-VAE-conf: conformalized QR-AVE; QR-VAE-GS: Gaussian QR-VAE; QR-VAE-GS-conf: Gaussian conformalized QR-VAE. }
\centering
\begin{tabular}{ |c|c|c|c|c|c|} 
\hline
& VAE & QR-VAE &  QR-VAE-conf & QR-VAE-GS& QR-VAE-GS-conf\\
\hline
AUC   & 0.52 & N/A & N/A &0.92 & 0.92 \\ 
\hline
Dice coefficient & 0.006 &0.32 &0.37 &0.37 &0.41\\
\hline
\end{tabular}
\label{tab:1}
\end{table}

\subsection{Supervised Lesion Detection}
\label{subsec:sup-lesion-detection}
We evaluated our BQR supervised approach on the LIDC-IDRI dataset \citep{armato2011lung}. This data consists of 1018 3D thorax CT scans annotated by four radiologists tasked with finding multiple lung nodules in each scan.  The data is ideal for capturing inherent uncertainty in the data labels that comes from disagreement between
experts. The data was preprocessed as described by \cite{kohl2018probabilistic}. They extracted 2D slices centred around the annotated
nodules and generated 180 x 180 images  when at least one expert has segmented a nodule. This process resulted in a dataset of 8882 images in the training set, 1996 images in the validation
set, and 1992 images in the test set. We reshaped the data into 128 x 128 images for input to the neural network. We used a U-Net architecture  \citep{ronneberger2015u} with 2D convolutional layers. The output layer was modified to generate four quantiles with four output branches using softmax activation. We compared the performance of BQR U-Net with the deterministic U-Net \citep{ronneberger2015u}. As ground truth for the comparison, first we estimated agreement maps for each test image by combining the lesion annotations of the four raters. This generates, for each image, an annotation with $P(Y=1|X)$ values greater than or equal to 0, 0.25, 0.5, 0.75,1. $X$ here represents the input image. Given an input image, the BQR U-Net generates output regions where probabilities of the label $Y=1$ are at or above the given quantile thresholds. The BQR U-Net was trained with the loss function in eq. \ref{eq:bqr_loss}. For comparison, the deterministic U-Net was trained using the binary cross-entropy loss. The BQR U-Net was trained to output
0.125, 0.375, 0.625, 0.875 quantiles. These quantile values represent the centroids of the intervals between the test data quantiles (0, 0.25, 0.5, 0.75,1) representing 0-4 rater agreements respectively. We used these thresholds rather than the same quantiles as the test data to avoid operating at the boundary points between operators resulting from combining data from four raters only. To generate the corresponding estimated quantiles for the deterministic U-Net we thresholded the softmax probability at 0.125, 0.375, 0.625, 0.875.

Both deep BQR and deterministic U-Net diverged to the trivial solution of predicting all labels as zero due to the extreme class imbalance when initialized with a cold start. We therefore warmed up both models using a weighted cross-entropy loss for one epoch weighting samples by $1\div166$, the ratio of the zero to one labels in the training set. We trained both models for 5 epochs. Our results show no significant improvement for deep BQR compared to the cross-entropy loss in terms of Dice coefficients for different agreement areas. The Dice coefficients between estimated probability areas for deterministic U-Net and BQR U-Net are plotted in Figure \ref{fig:BQR_hist} and summarized in Table 2. In the figure we show the distribution of the Dice coefficients between the BQR and ground truth quantiles, DT (deterministic U-Net) and ground truth, and also between DT and BQR. In some cases, particularly for higher quantiles, there was low agreement between human raters in the ground-truth test data. For example, 64 percent of the test data showed zero pixels in common between all four human raters, leaving the 0.875 quantile empty for those images. We therefore computed the Dice coefficients only over those regions in which the test data had a non-empty data set for that quantile. The results show reasonable agreement for the 0.125, 0.375 and 0.625 quantiles, but relatively poor results for the 0.875 quantile. Surprisingly, results for BRQ and the determinsic U-Net (DT U-Net) are very similar, even though the actual degree of overlap between the two is no better than between each of them and the ground truth labels. The fact that both methods perform poorly for the 0.875 quantile reflects both that the data are of relatively poor quality in this region and also that performance is likely limited by the imbalance in the training data between non-lesional areas and lesions confidently identified by all four raters.   
Here we used the deterministic U-Net as a backbone since it is arguably the most commonly used network for medical image segmentation task. Other networks could also be used as a backbone. To investigate whether we would expect further improvements using a probabilistic backbone, we also implemented the Probabilistic U-Net (Prob. U-Net) \citep{kohl2018probabilistic} to capture rater uncertainty. Our results show that for the quantile estimation task, the performance of Prob. U-Net and U-Net are comparable. As a result, it appears unlikely that replacing the U-Net with its proabilistic form in the backbone would lead to significant improvements.

\begin{figure}
\centering
\includegraphics[width=.8\textwidth]{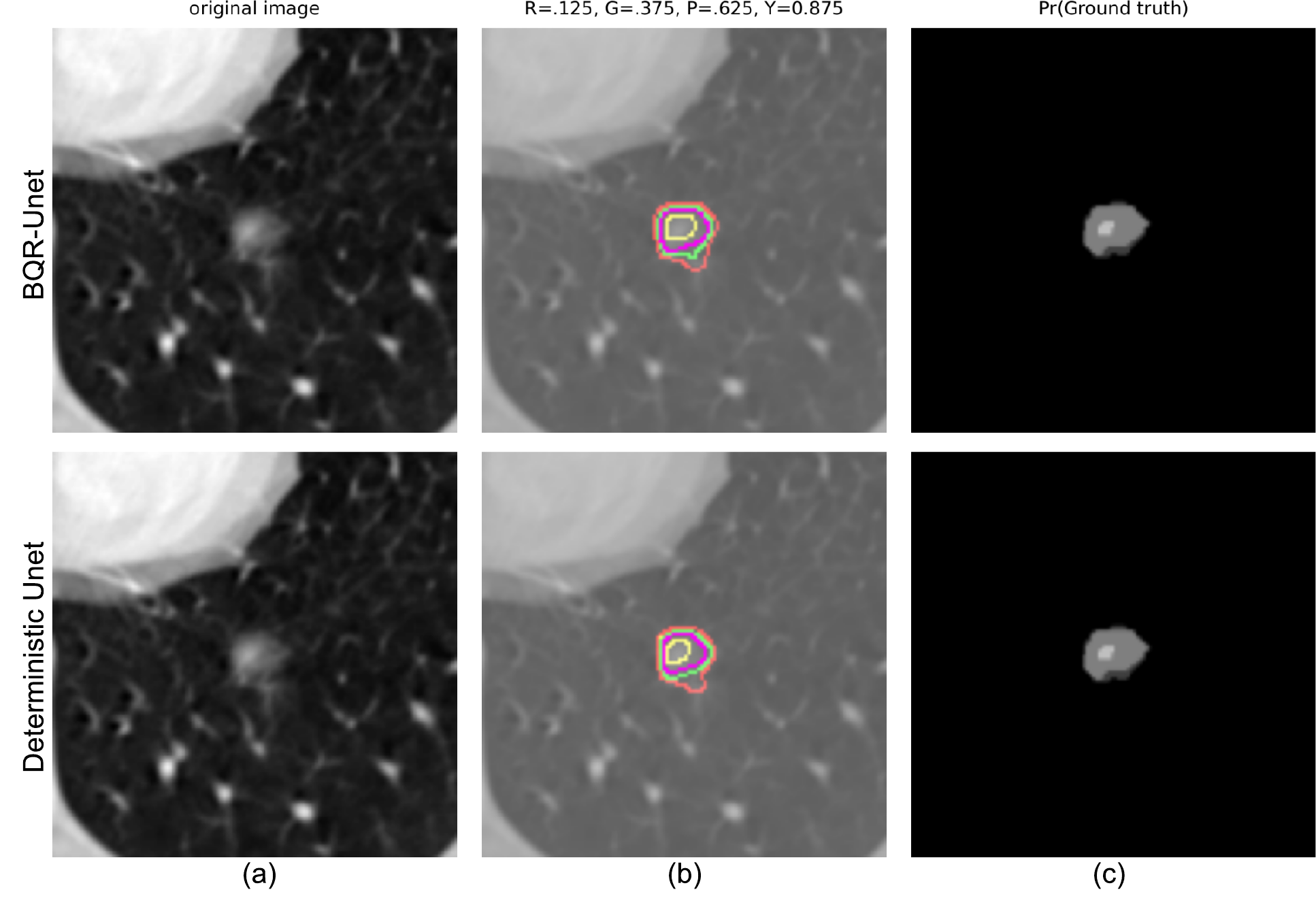}
\caption{Top row: results of U-Net delineation of lesion boundaries.  Bottom row: results of deterministic cross-entropy U-Net. (a) The original slice of the lung image; (b) estimated probability regions corresponding to 0.125,0.375,0.625,0.875 quantile levels shown with Red, green, purple and yellow colors respectively; (c) the estimate of thresholded lesion  boundary from human raters corresponding to agreement between 1,2,3 and 4 raters. 
}
\label{fig:BQR}
\end{figure}

\begin{figure}
\begin{center}
\includegraphics[width=1\textwidth]{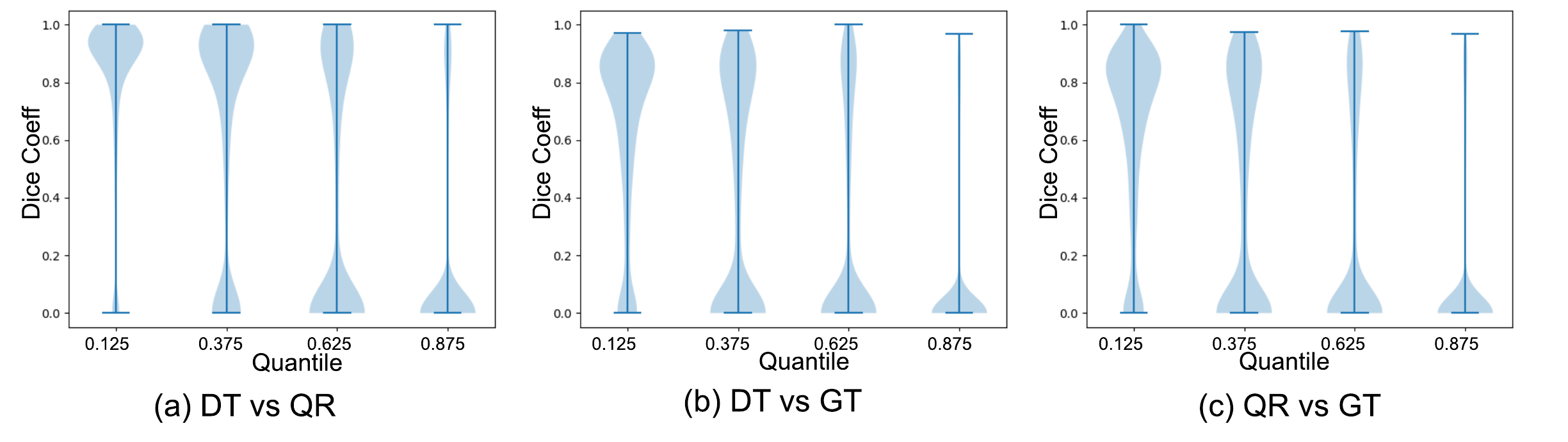}
\end{center}
\caption{Violin plots of the Dice coefficients between quantiles $(0.125, 0.375,0.625, 0.875)$) and rater agreement maps for the test datasets}, GT: ground truth, DT:  binary cross-entropy (Deterministic U-Net), QR: quantile regression (BQR U-Net). The fraction of empty quantiles in the ground truth (excluded from Dice coefficient computations) were $0.07, 0.31, 0.45, 0.64$ respectively. The width of the violin indicates the fraction of the dataset as a function of the dice coefficient.

\label{fig:BQR_hist}
\end{figure}

\begin{table}
\caption{The mean (std dev) of the Dice coefficients between estimated probability regions $(P(Y=1|X)\geq \alpha$, where  $\alpha$ is $(0.25, 0.5,0.75, 1)$; GT: ground truth, DT: deterministic, $P = P(Y=1|X)$}
\centering
\begin{tabular}{|c|c|c|c|c|c|} 
\hline
& $P\geq 0.25$& $P \geq 0.5$ & $P\geq 0.75$ &  $P=1$ \\
\hline
BQR U-Net vs GT  &  0.68(0.27)& 0.60(0.34) &0.50(0.40) &0.27(0.35)
 \\ 
\hline
DT U-Net vs GT&  0.67(0.27) & 0.59(0.34) & 0.50(0.38) & 0.32(0.37)\\
\hline
DT U-Net vs BQR U-Net & 0.81(0.27) &0.63(0.40) &0.40(0.43) &0.16(0.33)\\
\hline
\color{black}
Prob. U-Net vs GT & \color{black}0.60(0.26) &\color{black}0.60(0.39) &\color{black}0.51(0.43) &\color{black}0.31(0.36)\\

\hline
\end{tabular}
\label{tab:2}
\end{table}

\section{Conclusion}
Quantile regression is a simple yet powerful method for estimating uncertainty both in supervised and unsupervised lesion detection. We proposed novel cost functions to apply quantile regression and capture confidence intervals for lesion segmentation.

In the unsupervised framework we used the VAE, a popular model for unsupervised lesion detection \citep{chen2018unsupervised,baur2018deep,pawlowski2018unsupervised}. VAEs can be used to estimate reconstruction probability instead of reconstruction error for anomaly detection tasks. For calculating reconstruction probability, VAE models the output as a conditionally independent Gaussian characterized by means and variances for each output dimension. Simultaneous estimation of the mean and the variance in VAE underestimates the true variance leading to instabilities in optimization \citep{skafte2019reliable}. For this reason, classical VAE formulations that include both mean and variance estimates are rarely used in practice. Typically, only the mean is estimated with variance assumed constant \citep{skafte2019reliable}. To address this problem in variance estimation, we proposed an alternative quantile-regression model (QR-VAE) for improving the quality of variance estimation. We used quantile regression and leveraged the Guassian assumption to obtain the mean and variance by estimating two quantiles. We showed that our approach outperforms VAE as well as a Comb-VAE which is an alternative approach for addressing the same issue, in a synthetic as well as real world dataset. Our approach also has a more straightforward implementation compared to Comb-VAE. As a demonstrative application, we used our QR-VAE model to obtain a probabilistic heterogeneous threshold for a brain lesion detection task. This threshold results in a completely unsupervised lesion (or anomaly) detection method that avoids the need for a labeled validation dataset for principled selection of an appropriate threshold to control the false discovery rate. Beyond the current application we note that Quantile regression is applicable to deep learning models for medical imaging applications beyond the VAE and anomaly detection as the pinball loss function is easy to implement and optimize and can be applied to a broader range of network architectures and cost functions. 

For supervised lesion detection, we present deep binary quantile regression to estimate label uncertainty. Specifically, we use this technique to estimate quantiles of the labels that represent uncertainty. The lesion segmentations generated for each quantile reflect this uncertainty. Using LIDC data with 4 annotations we aimed to estimate the disagreement between the annotators.
Although it has been reported that deep binary QR has a better performance in imbalanced datasets in lesion segmentation task with extreme imbalance toward the class zero (normal), warming up the model was still needed in order to prevent it from converging to the trivial solution. Our result show no significant improvement in terms of dice coefficient between ground truth and estimated areas of agreement for deep binary QR compared to a deterministic U-Net. We found relatively small agreement for the 0.875 quantile region for these two estimations (row three, Table \ref{tab:2}) demonstrating that although we obtain similar performance, these two estimators are not annotating the same region regions. This finding indicates the potential for further improvements in the performance of both methods. Based on the current results, while numerical results are similar, the fact that with fewer training samples QR is less likely to diverge than the deterministic U-Net indicates that the QR approach may be more robust and stable.  

We investigated the advantages of using quantile regression in both supervised and unsupervised settings.  In the unsupervised framework, the estimated confidence interval is used to capture uncertainty from which can identify outliers that represent our detected lesions. We demonstrated  the advantage of using this quantile regression approach in the VAE setting. In the supervised framework, we used BQR to estimate the uncertainty of raters for the case where multi-rater data is available for training (and testing).

 

\section*{Acknowledgements}
This work was supported by NIH grants R01 NS074980, R01 NS089212, and R01 EB026299, and by the DOD grant W81XWH-18-1-061.

\bibliography{ref}

\end{document}